\documentclass[conference]{IEEEtran}
\IEEEoverridecommandlockouts
\usepackage{cite}
\usepackage{amsmath,amssymb,amsfonts}
\usepackage{algorithmic}
\usepackage{graphicx}
\usepackage{textcomp}
\usepackage{xcolor}
\usepackage{booktabs}
\usepackage{multirow}
\usepackage{subcaption} 
\usepackage[most]{tcolorbox} 

\def\BibTeX{{\rm B\kern-.05em{\sc i\kern-.025em b}\kern-.08em
    T\kern-.1667em\lower.7ex\hbox{E}\kern-.125emX}}
\begin{document}

\title{Benchmarking Machine Learning Models for IoT Malware Detection under Data Scarcity and Drift\\[-1mm]
{\normalsize Author pre-print copy. The manuscript has been accepted for publication at CCNC 2026}\\[-6mm]
{\normalsize © 2026 IEEE. Personal use of this material is permitted.}\\[-6mm]
{\normalsize Permission from IEEE must be obtained for all other uses, in any current or future media, including reprinting/republishing}\\[-6mm]{\normalsize this material for advertising or promotional purposes, creating new collective works, for resale or redistribution to servers or}\\[-6mm] {\normalsize lists, or reuse of any copyrighted component of this work in other works.}}

\author{
    \IEEEauthorblockN{
    Jake Lyon$^*$\textsuperscript{1},
    Ehsan Saeedizade$^*$\textsuperscript{2},
    Shamik Sengupta\textsuperscript{2}
    }
    \IEEEauthorblockA{
    \textsuperscript{1}Department of Computer Science, The College of Wooster, Wooster, USA \\
    \textsuperscript{2}Department of Computer Science and Engineering, University of Nevada-Reno, Reno, USA \\
    jlyon26@wooster.edu, esaeedizade@unr.edu, ssengupta@unr.edu
    }
}
\maketitle

\begingroup
\renewcommand\thefootnote{}\footnotetext{* Equal contributing authors}
\endgroup

\begin{abstract}
    The rapid expansion of the  Internet of Things (IoT)  in domains such as smart cities, transportation, and industrial systems has heightened the urgency of addressing their security vulnerabilities. IoT devices often operate under limited computational resources, lack robust physical safeguards, and are deployed in heterogeneous and dynamic networks, making them prime targets for cyberattacks and malware applications. Machine learning (ML) offers a promising approach to automated malware detection and classification, but practical deployment requires models that are both effective and lightweight. The goal of this study is to investigate the effectiveness of four supervised learning models (Random Forest, LightGBM, Logistic Regression, and a Multi-Layer Perceptron) for malware detection and classification using the IoT-23 dataset. We evaluate model performance in both binary and multiclass classification tasks, assess sensitivity to training data volume, and analyze temporal robustness to simulate deployment in evolving threat landscapes. Our results show that tree-based models achieve high accuracy and generalization, even with limited training data, while performance deteriorates over time as malware diversity increases. These findings underscore the importance of adaptive, resource-efficient ML models for securing IoT systems in real-world environments.

\end{abstract}

\begin{IEEEkeywords}
Malware Detection, Internet Of Things, Machine Learning, Cyber Security
\end{IEEEkeywords}

\section{Introduction}
\noindent The Internet of Things (IoT) is transforming domains such as smart transportation, industrial automation, and connected urban infrastructure. These decentralized networks consist of resource-constrained devices using diverse communication protocols, which increases flexibility but also introduces security vulnerabilities\cite{neshenko2019demystifying}. Limited computational resources, insecure protocols, and weak physical safeguards make IoT devices attractive targets for cyber threats such as botnets, denial-of-service (DoS) attacks, and malware infections \cite{vitorino2021comparative}.


Traditional malware detection systems based on signature matching and deep packet inspection are often too computationally intensive for IoT environments. As a result, machine learning (ML) approaches have emerged as promising alternatives, offering an automated, lightweight, and scalable alternative for detecting anomalies and malicious behaviors in network traffic \cite{vitorino2021comparative,saeedizade2023burst}. However, real-world deployment of ML-based approaches faces several critical challenges: scarcity of labeled data, class imbalance, and evolving attack patterns that reduce the generalizability of ML models.

This study systematically evaluates four supervised learning models, Logistic Regression (LR), Random Forest (RF), LightGBM (LGBM), and Multi-Layer Perceptron (MLP) on the IoT-23 dataset\cite{sebastian_garcia_2020_4743746}. Our investigation explores three key dimensions:

\begin{itemize}
    \item \textbf{Model performance in classification tasks:} We assess the ability of each model to detect and classify malware under both binary and multiclass settings. This comparison helps identify models that strike the best balance between accuracy, complexity, and suitability for deployment in resource-constrained environments.
    \item \textbf{Data efficiency and training size sensitivity:} Since labeled IoT traffic data can be difficult and expensive to obtain, understanding how model performance scales with training data availability is essential. We evaluate the impact of different training data sizes to determine how much labeled data is needed to achieve reliable performance, with particular attention to model robustness in multiclass scenarios where class imbalance can distort learning.
    \item \textbf{Temporal robustness and generalization over time:} IoT environments are dynamic, which means malware behavior, device activity, and network configurations can shift over time. This means that a model trained on historical traffic may degrade when exposed to emerging threats. To simulate such deployment scenarios, we perform a temporal analysis using rolling time windows and evaluate performance drift over successive months. This offers insights into the different models’ resilience to concept drift and the need for continual retraining.
\end{itemize}

By examining these dimensions, this study provides a practical and comprehensive analysis of the strengths and limitations of supervised ML models for malware detection in dynamic IoT environments. Our findings contribute to the design of lightweight, scalable, and temporally adaptive detection systems suitable for real-world IoT deployments.

The rest of the paper is organized as follows: The rest of the paper is organized as follows: Section 2 reviews related work on ML-based IoT malware detection. Section 3 details the methodology, including dataset preparation, feature engineering, and evaluation procedures. Section 4 presents results from the classification, data efficiency, and temporal robustness analysis. Section 6 concludes the paper.

\section{Related Works}

\noindent Recent studies have evaluated the applicability of various ML models for IoT malware detection. Jamal et al.\cite{jamal2022malware} employed a three-layer artificial neural network on the TON IoT dataset and achieved classification and detection accuracies of $0.9708$ and $0.9417$, respectively.  Zi Wei et al.\cite{zi_wei_comparing_2023} compared Random Forest and Naïve Bayes models using the IoT-23 dataset, concluding that Random Forest consistently outperformed Naïve Bayes across precision, recall, and F1 score metrics.  Vitorino et al. \cite{vitorino2021comparative} conducted a broader comparative analysis of various ML models, such as Support Vector Machines (SVM), XGBoost, LightGBM, Isolation Forest (iForest), Local Outlier Factor (LOF), and Deep Reinforcement Learning (DRL), also using the IoT-23 dataset. Their results showed LightGBM achieving the best overall classification performance, while iForest excelled in anomaly detection, and DRL offered advantages in dynamic learning environments by enabling continuous learning.

Song et al. provided a systematic review of deep learning applications in malware detection, covering CNNs, RNNs, and GANs f\cite{song2025application}. Ferdous et al. presented a survey on ML-based malware detection across multiple platforms, including PCs, mobile devices, IoT, and cloud environments \cite{ferdous2025survey}. Bensaoud et al. conducted a survey dedicated to deep learning approaches for malware detection across major operating systems \cite{bensaoud2024survey}. 


Prior studies have demonstrated the promise of ML and DL for malware detection, but often focus on improving performance metrics such as accuracy or broad surveys without addressing deployment challenges. Our study extends this line of work by systematically analyzing supervised ML models for IoT malware detection. We examine model performance in both binary and multiclass classification tasks, assess data efficiency, and analyze the temporal robustness of ML models. This broader evaluation offers a deployment-aware perspective on ML-based malware detection in dynamic IoT networks.

\section{Methodology}
\begin{figure}
    \centering
    \includegraphics[width=0.75\linewidth]{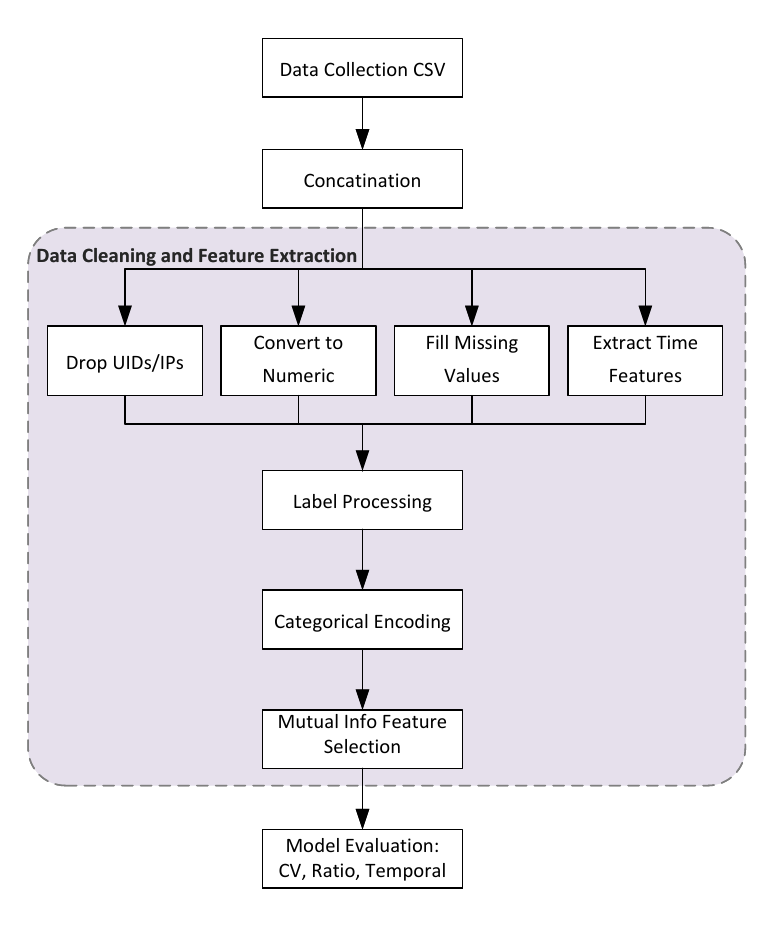}
    \caption{Methodology workflow}
    \label{fig:Methodology_Workflow}
    \vspace{-5mm}
\end{figure}

\noindent This section outlines the dataset used in the study, the preprocessing steps applied to prepare the data for analysis, and the evaluation strategies employed to assess the performance and robustness of the selected machine learning models. The overall workflow is summarized in Fig.\ref{fig:Methodology_Workflow}.

\subsection{Data Collection}
\noindent Network traffic data can be collected using tools such as Wireshark \cite{wireshark_guide} or tcpdump\cite{tcpdump_homepage}. These tools capture real-time raw packets and store them in Packet Capture (PCAP) format. PCAP files include low-level details such as protocol headers, payload content, and communication protocols like TCP, UDP, and ICMP. In this study, we utilized the well-known and publicly available IoT-23 dataset \cite{sebastian_garcia_2020_4743746}, which contains labeled network flows derived from PCAP files. This dataset includes both benign and malicious traffic samples across a variety of malware families and attack types, making it well-suited for evaluating ML-based malware detection approaches.

\subsection{Data Preprocessing}
\noindent Collected raw PCAP files can not be used directly by machine learning models. Therefore, tools such as CICFlowMeter\cite{cicflowmeter_github} and Zeek\cite{zeek_website} are typically employed to extract flow-level features from PCAP files to tabular formats by aggregating metrics such as packet counts, durations, byte rates, and protocol types. The dataset that is used in this study provides extracted features in multiple CSV files \cite{iot23kaggle}. We apply the following additional preprocessing steps to prepare the data for analysis:

\begin{itemize}
    \item \textbf{Concatenation:} The dataset is originally provided in separate CSV files corresponding to different capture sessions. These files are concatenated into a single unified dataset to enable consistent training and evaluation across all samples.
    \item \textbf{Cleaning:} Irrelevant, missing, or session-specific fields such as connection IDs and IP addresses are removed to reduce noise and avoid data leakage.
    \item \textbf{Timestamp Engineering:} Raw timestamps are used to derive two contextual features as $day-of-week$ and $hour-of-day$ for training. Timestamp data also enables segmentation of the dataset into monthly intervals for temporal evaluation.
    \item \textbf{Feature Encoding and Labeling:} To train machine learning models, categorical fields (e.g., $protocol~types$, $connection~state$) are encoded using label or one-hot encoding. The original string-based labels are transformed into two formats: a binary label (benign vs. malicious) and a multiclass label (specific malware families). This transformation enables models to learn from categorical patterns without misinterpreting string-based formats
    \item \textbf{Feature Selection:} To reduce dimensionality and computational overhead, we apply mutual information gain analysis \cite{scikit_mutual_info_classif} to quantify the relevance of each feature to the target label.
    \item \textbf{Feature Scaling:}  All numerical features are normalized using Min-Max scaling to the $[0,1]$ range.
\end{itemize}


\subsection{Model Training and Evaluation}
\noindent We evaluate the performance of four supervised learning models: Logistic Regression, Random Forest, LightGBM, and a Multi-Layer Perceptron. 
We tune each model's hyperparameters using RandomizedSearchCV with 50 random configurations and a fixed random seed for reproducibility. The specific architectures and tuned hyperparameters of each model are discussed in the section \ref{sec:evaluation}. The evaluation framework consists of three distinct analyses:

\begin{itemize}
    \item Performance evaluation: We perform 5-fold cross-validation to assess model performance on binary and multiclass malware detection tasks.
    \item Training Size Sensitivity: We analyze the effect of training data availability by training models on varying proportions of the dataset (from $30\%$ to $80\%$) and observe how their performance changes with data availability.
    \item Temporal Evaluation: We adopt a rolling-window approach, training models on earlier months and testing on later ones to simulate real-world deployment. This setup evaluates how models generalize to evolving attack patterns and temporal drift.
\end{itemize}

To ensure consistency, we reuse the tuned hyperparameters from the cross-validation phase when evaluating models under training size and temporal conditions, as the focus is on assessing model behavior rather than maximizing performance.

\section{Evaluation and Analysis}
\label{sec:evaluation}
\noindent This section provides the evaluation results. We first outline the experimental setup, then present results for each analysis.

\subsection{Experimental Setup}
\subsubsection{Dataset}
All experiments are conducted using the IoT-23 dataset, a widely-used benchmark for evaluating network intrusion detection in IoT environments \cite{sebastian_garcia_2020_4743746, iot23kaggle}. The dataset contains 23 labeled network traffic captures in pcap format, 20 malware-infected and 3 benign, collected between 2018 and 2019 by the Stratosphere Laboratory. The dataset includes over 760 million packets and 325 million labeled flows, covering diverse attack behaviors such as command-and-control (C\&C) communications, distributed denial of service (DDoS), horizontal port scans, and malicious file downloads. 

\begin{table}[ht]
\caption{Selected features based on mutual information score analysis}
\label{tab:selected-features}
\centering
\begin{tabular}{l p{5cm}}  
\toprule
\textbf{Feature} & \textbf{Description} \\
\midrule
\midrule
\textbf{id.orig\_p} & Source TCP/UDP port (/ ICMP code) \\
\textbf{id.resp\_p} & Destination TCP/UDP port \\
\textbf{proto} & Transport layer protocol \\
\textbf{duration} & Period of connection \\
\textbf{orig\_bytes} & Source payload bytes; from sequence number if TCP \\
\textbf{conn\_state} & Connection state \\
\textbf{history} & Connection state history \\
\textbf{orig\_pkts} & Number of packets from source \\
\textbf{orig\_ip\_bytes} & Number of source IP bytes \\
\textbf{resp\_ip\_bytes} & Number of destination IP bytes \\
\textbf{hour} & Hour of the day \\
\textbf{weekday} & Day of the week \\
\bottomrule
\end{tabular}
\vspace{-3mm}
\end{table}

\begin{figure}
    \centering
    \includegraphics[width=0.8\linewidth]{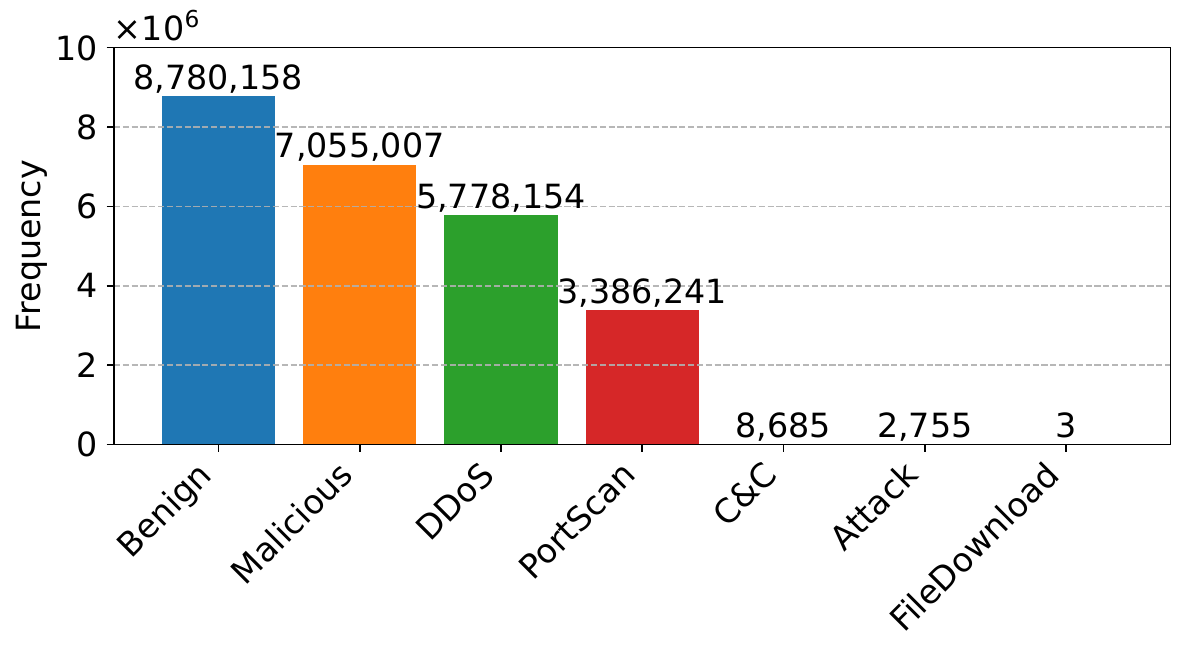}
    \caption{Label distribution of the sampled dataset}
    \label{fig:malware_frequency_bar_chart}
    \vspace{-3mm}
\end{figure}

In this paper, we use a processed version of I0T-23 that provides flow-level features in CSV format \cite{iot23kaggle}. This dataset included $21$ features, of which 5 were removed during preprocessing. Specifically, the unique connection identifier ($uid$) was discarded as it does not contribute meaningful information for classification. The source and destination IP address fields ($id.orig\_p$ and $id.resp\_h$) were excluded as IP addresses vary heavily. Additionally, the $local\_orig$ and $local\_resp$ features were dropped because they contained no recorded values across the dataset. For performance evaluation, we sample $25$ million flows due to hardware constraints for model training. The label distribution in the sampled dataset is presented  in Figure \ref{fig:malware_frequency_bar_chart}.

\subsubsection{Feature Selection}
Following initial cleaning, we apply filter-based feature selection using mutual information gain to reduce dimensionality and focus on the most informative attributes. Table \ref{tab:selected-features} presents the top 10 selected features by this process, along with the two time-related contextual features: day of the week and hour of the day.

\begin{table*}[!ht]
\centering
\caption{Tuned hyperparameters for machine learning models}
\label{tab:hyperparams}
\begin{tabular}{l l p{0.65\textwidth}}
\toprule
\textbf{Model} & \textbf{Task} & \textbf{Hyperparameters} \\
\midrule
\multirow{2}{*}{Logistic Regression} 
& Binary & \texttt{C=10}, \texttt{solver=saga}, \texttt{penalty=elasticnet}, \texttt{l1\_ratio=1}, \texttt{max\_iter=1000}
\\
\addlinespace[2pt]
\cline{2-3}
\addlinespace[2pt]
& Multiclass & \texttt{C=0.1},  \texttt{solver=saga}, \texttt{penalty=elasticnet}, \texttt{l1\_ratio=1}, \texttt{max\_iter=1000}
\\
\midrule
\multirow{2}{*}{Random Forest} 
& Binary & \texttt{n\_estimators=200}, \texttt{max\_depth=20}, \texttt{min\_samples\_split=5}, \texttt{min\_samples\_leaf=1}, \texttt{max\_features=log2}
\\
\addlinespace[2pt]
\cline{2-3}
\addlinespace[2pt]
& Multiclass & \texttt{n\_estimators=100}, \texttt{max\_depth=20}, \texttt{min\_samples\_split=5}, \texttt{min\_samples\_leaf=1}, \texttt{max\_features=sqrt}
\\
\midrule
\multirow{2}{*}{LightGBM} 
& Binary & \texttt{objective=binary}, \texttt{num\_leaves=127}, \texttt{learning\_rate=0.05}, \texttt{n\_estimators=500},  \texttt{colsample\_bytree=0.6}, \texttt{max\_depth=10}, \texttt{subsample=0.8}
\\
\addlinespace[2pt]
\cline{2-3}
\addlinespace[2pt]
& Multiclass & \texttt{objective=multiclass},  \texttt{num\_leaves=63}, \texttt{learning\_rate=0.01}, \texttt{n\_estimators=500},  \texttt{colsample\_bytree=0.8}, \texttt{max\_depth=5}, \texttt{subsample=1.0}, \text{epochs=50}, \texttt{batch\_size=256}
\\
\midrule
\multirow{2}{*}{MLP (Deep Learning)} 
& Binary & \texttt{optimizer=Adam}, \texttt{learning\_rate=0.0001}, \texttt{loss=binary\_crossentropy}, \texttt{metrics=accuracy}, \texttt{activation=sigmoid}, \texttt{epochs=50}, \texttt{batch\_size=256}
\\
\addlinespace[2pt]
\cline{2-3}
\addlinespace[2pt]
& Multiclass & \texttt{optimizer=Adam}, \texttt{learning\_rate=0.0001}, \texttt{loss=sparse\_categorical\_crossentropy}, \texttt{metrics=accuracy}, \texttt{activation=softmax}, \texttt{epochs=50}, \texttt{batch\_size=256}
\\
\bottomrule
\end{tabular}
\vspace{-3mm}
\end{table*}

\subsubsection{ML models}
We evaluated four supervised learning models, selected based on their popularity and prior success in network intrusion detection tasks \cite{hamid2016machine}. All models were implemented using the scikit-learn and Keras/TensorFlow libraries in Python. The tuned hyperparameters used for both tasks across all models are summarized in Table~\ref{tab:hyperparams}.

\begin{enumerate}
    \item \textbf{Logistic Regression (LR)} is a linear model often used as a lightweight baseline. Its simplicity makes it attractive for constrained environments, but its linear nature limits its capacity to capture complex decision boundaries \cite{rahman2023cognitive}. 
        
    \item \textbf{Random Forest (RF)} is an ensemble of decision trees trademarked by Leo Brieman and Adele Cutler, that improves predictive power by reducing overfitting and variance. It is widely adopted in cybersecurity for its robustness and low sensitivity to parameter tuning. The algorithm utilizes the output of multiple decision trees to reach a single result. This often produces more accurate results, and the use of an ensemble helps reduce variance within a noisy dataset\cite{SupervisedLearningModels}. 

    \item \textbf{LightGBM (LGBM)} uses an ensemble of decision trees and is a gradient boosting framework. LightGBM uses  Gradient-based One-Side Sampling (GOSS) to build decision trees and utilizes Exclusive Feature Bundling (EFB) to reduce the number of features and improve the training efficiency. Its performance in intrusion detection tasks has been highlighted in several prior studies\cite{ke2017lightgbm}.  
    
    \item \textbf{Multi-Layer Perceptron (MLP)} is used as a feedforward neural network and can model complex, non-linear relationships in the data \cite{hamid2016machine}. The MLP architecture consists of an input layer, two hidden layers with $64$ and $128$ neurons respectively, and an output layer with $softmax$ activation for multiclass classification and $sigmoid$ for binary classification. Each hidden layer uses $ReLU$ activation, and a $0.2$ dropout rate is applied after each hidden layer to prevent overfitting.
\end{enumerate}

\subsubsection{Performance Metrics}
\noindent We use the following common evaluation metrics in the literature to evaluate the performance of ML models \cite{caballar2025model_performance}:
\begin{itemize}
    \item \textbf{Accuracy} is calculated as the proportion of correct predictions over all instances. In this study, this represents the percentage of correctly classified network traffic.
    \item \textbf{Precision} is the proportion of true positives among predicted positives and is calculated as $Precision = \frac{TP}{TP+FP}$. 
    \item \textbf{Recall} is the metric that quantifies the number of true positives among actual positives and is calculated as $Recall = \frac{TP}{TP+FN}$.  It is known as the sensitivity and is important for detecting all malicious activity.
    \item \textbf{F1-Score} is the harmonic mean of precision and recall and is calculated as $F=\frac{2 \cdot P \cdot R}{P+R}$ where $P$ and $R$ are precision and recall, respectively. This metric balances the trade-off between false positives and false negatives and is particularly important under class imbalance.
\end{itemize}

\subsection{Performance Evaluation}
\noindent We assess the model performance in two different tasks. In the binary classification task setting, models are trained to perform malware detection and classify data into malicious vs. benign classes. As can be seen in Table \ref{tab:BinaryPerformance}, all models achieve strong results in this task. RF and LGBM slightly outperform the others, both attaining F1 scores above $0.998$. While LR achieves the least performance and it reached a solid F1 of $0.9799$, demonstrating its viability in low-complexity detection scenarios.

\begin{table*}[ht]
\centering
\caption{Model performance by training ratio  in binary classification}
\label{tab:BinaryRatioTwoTiered}
\resizebox{0.95\linewidth}{!}{%
\begin{tabular}{c cccc|cccc|cccc|cccc}
\toprule
\textbf{Train Ratio} &
\multicolumn{4}{c}{\textbf{LR}} &
\multicolumn{4}{c}{\textbf{RF}} &
\multicolumn{4}{c}{\textbf{LGBM}} &
\multicolumn{4}{c}{\textbf{MLP}} \\
\cline{2-17}
& Accuracy (\%)& F1 Score & Precision & Recall &
  Accuracy (\%)& F1 Score & Precision & Recall &
  Accuracy (\%)& F1 Score & Precision & Recall &
  Accuracy (\%)& F1 Score & Precision & Recall \\
\hline

0.3 
& 98.11 &  0.9853 & 0.9974 &0.9734 
& 99..83 & 0.9987 & 0.9987 & 0.9979 
& 99.83 & 0.9987 & 0.9999 & 0.9975  
& 99.71 & 0.9977 & 0.9993 & 0.9962 
\\

0.4 
& 98.12 &  0.9853 & 0.9977 & 0.9734 
& 99.85 & 0.9989 & 0.9998 & 0.9980
& 99.83 & 0.9987 & 0.9999 & 0.9975
& 99.70 & 0.9977 & 0.9992 & 0.9962
\\

0.5 
& 98.11 & 0.9852 & 0.9973 & 0.9734 
& 99.87 & 0.9990 & 0.9998 & 0.9981
& 99.83 & 0.9987 & 0.9999 & 0.9975
& 99.70 & 0.9977 & 0.9992 & 0.9962
\\

0.6 
& 98.12 & 0.9854 & 0.9976 & 0.9734
& 99.88 & 0.9991 & 0.9999 & 0.9982
& 99.83 & 0.9987 & 0.9999 & 0.9975
& 99.70 & 0.9977 & 0.9993 & 0.9961
\\

0.7 
& 98.11 & 0.9853 & 0.9975 & 0.9734 
& 99.89 & 0.9991 & 0.9999 & 0.9984 
& 99.83 & 0.9987 & 0.9999 & 0.9975
& 99.71 & 0.9977 & 0.9993 & 0.9961
\\

0.8 
& 98.12 & 0.9854 & 0.9977 & 0.9734 
& 99.90 & 0.9992 & 0.9999 & 0.9985
& 99.83 & 0.9987 & 0.9999 & 0.9975
& 99.71 & 0.9968 & 0.9961 & 0.9975
\\

\bottomrule
\end{tabular}
}
\end{table*}

\begin{table*}[ht]
\centering
\caption{Model performance by training ratio  in multi-class classification}
\label{tab:multiRatioTwoTiered}
\resizebox{0.95\linewidth}{!}{%
\begin{tabular}{c cccc|cccc|cccc|cccc}
\toprule
\textbf{Train Ratio} &
\multicolumn{4}{c}{\textbf{LR}} &
\multicolumn{4}{c}{\textbf{RF}} &
\multicolumn{4}{c}{\textbf{LGBM}} &
\multicolumn{4}{c}{\textbf{MLP}} \\
\cline{2-17}
& Accuracy (\%)& F1 Score & Precision & Recall &
  Accuracy (\%)& F1 Score & Precision & Recall &
  Accuracy (\%)& F1 Score & Precision & Recall &
  Accuracy (\%)& F1 Score & Precision & Recall \\
\hline

0.3 
& 96.80 & 0.6125 & 0.5889 & 0.9512 
& 99.82 & 0.8372 & 0.8405 & 0.8341 
& 98.64 & 0.5648 & 0.5647 & 0.5652  
& 99.54 & 0.7227 & 0.7924 & 0.7291 
\\

0.4 
& 96.96 & 0.6087 & 0.5876 & 0.9510
& 99.84 & 0.8380 & 0.8381 & 0.8380
& 95.02 & 0.5439 & 0.5428 & 0.5471
& 99.59 & 0.7934 & 0.8109 & 0.7806
\\

0.5 
& 97.05 & 0.6111 & 0.5895 & 0.9532
& 99.86 & 0.9797 & 0.9801 & 0.9794
& 74.85 & 0.4143 & 0.4153 & 0.4161
& 99.26 & 0.7865 & 0.8106 & 0.7673
\\

0.6 
& 96.21 & 0.6064 & 0.5828 & 0.9495 
& 99.87 & 0.8364 & 0.8360 & 0.8370
& 72.02 & 0.4111 & 0.4356 & 0.4447
& 99.27 & 0.7942 & 0.8120 & 0.7791
\\

0.7 
& 96.95 & 0.6129 & 0.5902 & 0.9509
& 99.88 & 0.9792 & 0.9771 & 0.9816
& 87.59 & 0.5169 & 0.5224 & 0.5184
& 99.26 &  0.8032 & 0.8032 & 0.8124
\\

0.8 
& 96.36 & 0.6052 & 0.5830 & 0.9511
& 99.89 & 0.9769 & 0.9735 & 0.9808
& 97.70 & 0.5723 & 0.5712 & 0.61665
& 99.51 & 0.8005 & 0.8230 & 0.7854
\\

\bottomrule
\end{tabular}
}
\vspace{-3mm}
\end{table*}

\begin{table}[ht]
\centering
\caption{Model performance on binary classification}
\label{tab:BinaryPerformance}
\resizebox{0.8\linewidth}{!}{%
\begin{tabular}{lcccccc}
\toprule
\textbf{Model}  & \textbf{Accuracy (\%)} & \textbf{F1 Score} & \textbf{Precision} & \textbf{Recall} \\
\midrule
LR & 98.15 & 0.9799 & 0.9756 & 0.9848 \\
RF & \textbf{99.83} & \textbf{0.9981} & \textbf{0.9976} & \textbf{0.9986} \\
LGBM & \textbf{99.84} & \textbf{0.9982} & \textbf{0.9977} & \textbf{0.9987} \\
MLP & 99.67 & 0.9964 & 0.9957 & 0.9970 \\
\bottomrule
\end{tabular}
}
\end{table}%

\begin{table}[ht]
\centering
\caption{Model performance on multiclass classification}
\label{tab:MulticlassPeformance}
\resizebox{0.8\linewidth}{!}{%
\begin{tabular}{lcccc}
\toprule
\textbf{Model} & \textbf{Accuracy} & \textbf{F1 Score} & \textbf{Precision} & \textbf{Recall} \\
\midrule
LR  & 92.07 & 0.6571 & 0.6507 & 0.8322 \\
RF        & \textbf{99.81} & \textbf{0.9771} & \textbf{0.9916} & \textbf{0.9644} \\
LGBM             & \textbf{99.82} & \textbf{0.9421} & \textbf{0.9385} & \textbf{0.9482} \\
MLP       & 99.54 & 0.8531 & 0.9368 & 0.8508 \\
\bottomrule
\end{tabular}
}
\vspace{-3mm}
\end{table}

In the multiclass classification task, our goal is not only to detect the malicious class, but also to distinguish between specific malware families. This task is important in cases where different actions are needed in real-world scenarios to stop the malware. As shown in \ref{tab:MulticlassPeformance}, the performance diverges more significantly. However, RF again stand out with an F1 score of $0.9771$, indicating effective classification across a wide range of attack types. LGBM achieves similar accuracy but a lower F1-score ($0.9421$), suggesting some inconsistency, likely related to its difficulty in capturing minority class patterns. The MLP model yielded high precision but lower recall, which reduces its F1 score accordingly. Finally, LR has the weakest performance, with an F1 of $0.6571$, underscoring its limitations in handling complex class distributions.

\begin{tcolorbox}[
    colback=green!5!white, 
    colframe=green!50!black, 
    title=Takeaway, 
    width=0.9\linewidth,
    enlarge left by=0.02\linewidth, 
    box align=center
]
In IoT malware detection, tree-based models provide consistently strong performance in both binary and multiclass classification. MLP is a viable alternative when capturing complex patterns in a multi-class classification task is essential. LR, while less effective in multiclass scenarios, remains a suitable lightweight baseline for binary detection tasks.
\end{tcolorbox}

\subsection{Training Size Analysis}

In this section, we present the evaluation results for all models across training set sizes ranging from $30\%$ to $80\%$. Given the dataset’s class imbalance, we use the F1 score as the primary performance metric as it reflects the model's performance in terms of precision and recall.

As shown in Table~\ref{tab:BinaryRatioTwoTiered}, all models maintain consistently high performance across all training sizes in the binary classification task. This outcome is expected, as distinguishing between benign and malicious traffic is relatively straightforward, and even limited training sets provide sufficient signal for effective learning. As depicted, RF achieves the highest F1 score, reaching $0.9992$ with $80\%$ of the training data and already performing at $0.9987$ with only $30\%$. LightGBM closely follows with an F1 score of $0.9987$ at all levels, indicating remarkable stability. 

\begin{figure}[htbp]
    \centering
    \begin{subfigure}[a]{0.8\linewidth}
        \centering
        \includegraphics[width=\linewidth]{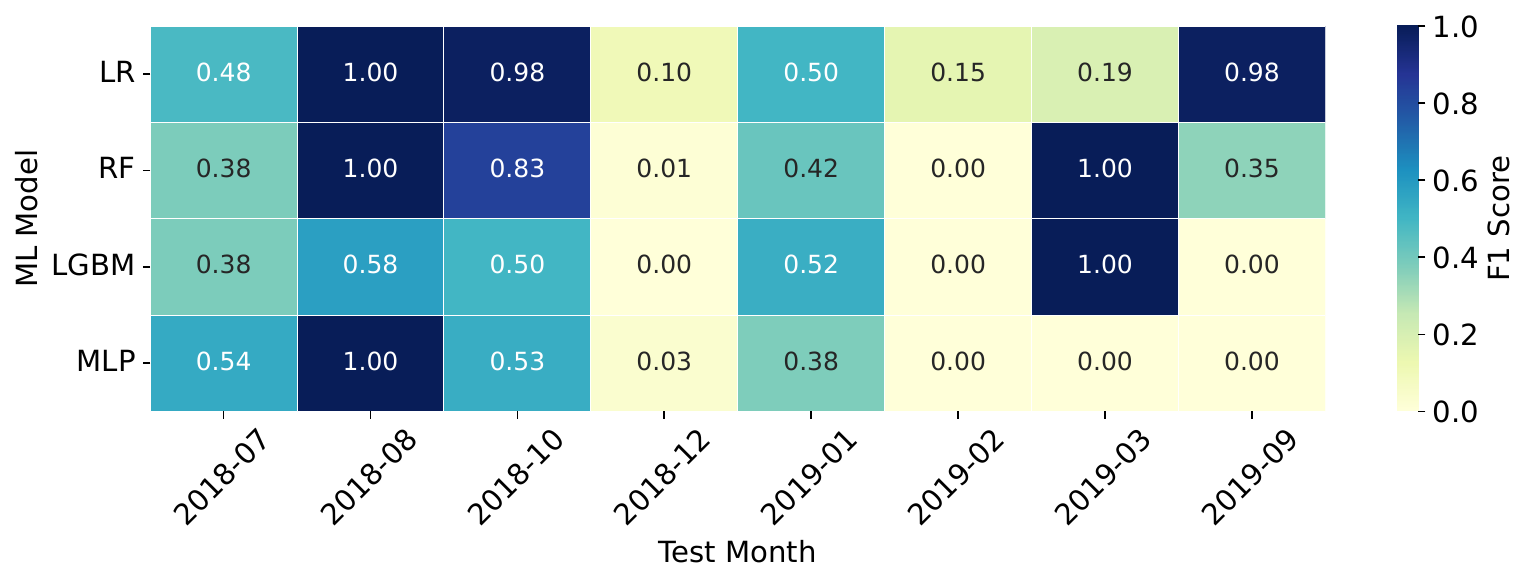}
        \caption{Binary classification}
        \label{fig:temp_bin}
    \end{subfigure}
    \hfill
    \begin{subfigure}[b]{0.8\linewidth}
        \centering
        \includegraphics[width=\linewidth]{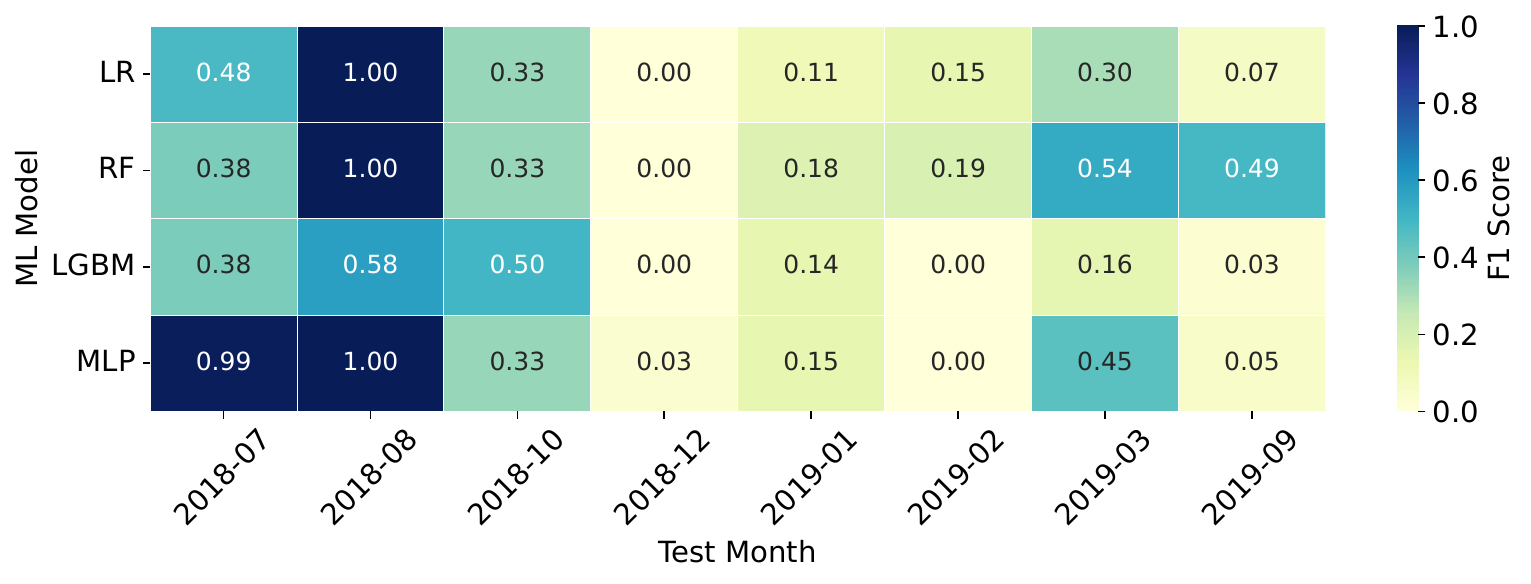}
        \caption{Multiclass classification}
        \label{fig:temp_multi}
    \end{subfigure}
    \caption{Temporal evaluation of F1 scores using a rolling-window setup, where models are trained on earlier months and tested on later ones, for (a) binary classification and (b) multiclass classification tasks across test months.}
    \label{fig:temp_all}
    \vspace{-5mm}
\end{figure}

However, as can be seen in Table~\ref{tab:multiRatioTwoTiered}, the multiclass classification results show more significant differences between models. RF again demonstrates outstanding performance, achieving an F1 score of $0.9797$ at just $50\%$ training size and maintaining similar results at higher ratios. RF also achieves remarkably higher Precision and Recall scores compared to the other models, indicating strong handling of class imbalance. MLP also shows a strong upward trend, with its F1 score improving from $0.7227$ at $30\%$ to $0.8005$ at $80\%$, indicating that additional data helps the neural model better capture complex decision boundaries. 

In Contrast, LR maintains a stable but significantly lower F1 score compared to the other models. While it achieved high recall (above $0.95$), its precision remained poor, reflecting the model's tendency to be more sensitive by overpredicting minority classes and thus increasing the false positive rate. LGBM, while achieving high accuracy (up to $97.7\%$ at $80\%$), suffers from poor F1 scores. The difference between accuracy and F1 score is mainly because of the class imbalance effect and highlights the model’s struggle to correctly classify minority classes.

\begin{tcolorbox}[
    colback=green!5!white, 
    colframe=green!50!black, 
    title=Takeaway, 
    width=0.96\linewidth,
    enlarge left by=0.02\linewidth, 
    box align=center
]
While all models perform well in binary classification, even with reduced data, RF consistently leads across tasks and training sizes. MLP improves with more data, while LR and LGBM struggle in multiclass settings due to class imbalance.
\end{tcolorbox}

\begin{table}[ht]
\centering
\caption{Monthly malware family distribution}
\resizebox{0.9\linewidth}{!}{%
\begin{tabular}{cp{7cm}}
\hline
\textbf{Month} & \textbf{Malware Families Present} \\
\hline
2018-05 & Benign, Malicious \\
2018-07 & Benign, Malicious \\
2018-08 & Benign, Malicious \\
2018-10 & Benign, Malicious C\&C \\
2018-12 & Benign, Malicious Attack, Malicious C\&C, Malicious DDoS, Malicious PartOfAHorizontalPortScan \\
2019-01 & Benign, Malicious C\&C, Malicious DDoS, Malicious FileDownload \\
2019-02 & Benign, Malicious Attack, Malicious C\&C, Malicious PartOfAHorizontalPortScan \\
2019-03 & Benign, Malicious Attack, Malicious C\&C, Malicious PartOfAHorizontalPortScan \\
2019-09 & Benign, Malicious C\&C, Malicious DDoS \\
\hline
\end{tabular}
}
\label{tab:malware-month-distribution}
\vspace{-3mm}
\end{table}

\subsection{Temporal Analysis}
Table~\ref{tab:malware-month-distribution} shows the malware family distribution across the months used in the analysis. As illustrated, the dataset begins with relatively simple malicious traffic in mid-2018 and gradually evolves to include more diverse and complex attacks. This highlights that ML-based malware detection approaches must adapt to evolving malware and attack patterns. To simulate the evolving conditions and evaluate the temporal robustness of each model, we use a rolling-window strategy. In this setup, models are trained on data from earlier months and evaluated on traffic from subsequent months, reflecting conditions where models encounter previously unseen attack types or distributions over time.

Fig.\ref{fig:temp_bin} shows the binary classification results. As can be seen, models perform well during the earlier months. This is because the diversity of malware families is low. As new malware types appear in later months, performance begins to degrade. Among all models, RF has a more stable performance over time. The performance gap widens in the multiclass classification task, as shown in Fig. \ref{fig:temp_multi}. As the number and complexity of classes increase, most models experience more pronounced degradation. RF again achieves a higher F2 score, reflecting a better precision and recall even in later months with multiple overlapping attack types.
\begin{tcolorbox}[
    colback=green!5!white, 
    colframe=green!50!black, 
    title=Takeaway, 
    width=0.96\linewidth,
    enlarge left by=0.02\linewidth, 
    box align=center
]
As cybersecurity threats diversity changes over time, even high-performing models trained on large volumes of historical data may struggle to maintain effectiveness over time without periodic retraining or adaptation. Therefore, to ensure sustained performance in dynamic IoT environments and address temporal drift, it is crucial to implement adaptive learning strategies, such as model fine-tuning, online learning, or active learning based on feedback from newly observed data.

\end{tcolorbox}

\section{Conclusion And Discussion}
This study provides a comprehensive evaluation of four supervised learning models for IoT malware detection under varying data availability and temporal conditions. Across binary and multiclass tasks, Random Forest and LightGBM delivered consistently high performance, while Logistic Regression served as a lightweight baseline, and MLP benefited from larger training sets. However, all models exhibited performance degradation over time, highlighting the impact of evolving malware patterns and the need for periodic retraining or adaptive learning strategies. These results emphasize that effective IoT security requires models that balance accuracy, efficiency, and adaptability to maintain robust performance in dynamic environments. In future work, we plan to design and evaluate adaptive learning techniques to address temporal drift and improve long-term robustness.

\section*{Acknowledgments}
The work in this study was supported by the NSF grants 2349092 and 2346755.

\bibliographystyle{IEEEtran}
\bibliography{main}

\end{document}